# TinyProp - Adaptive Sparse Backpropagation for Efficient TinyML On-device Learning

Marcus Rueb
Software Solutions / Artificial intelligence
Hahn-Schickard
Villingen-Schwenningen, Germany
Marcus.rueb@hahn-schickard.de

Daniel Maier
Software Solutions / Artificial intelligence
Hahn-Schickard
Villingen-Schwenningen, Germany
Daniel.Maier@hahn-schickard.de

Daniel Mueller-Gritschneder
Electronic Design Automation
TUM, Technical University Munich
Munich, Germany
daniel.mueller@tum.de

Axel Sikora
EMI
University of Applied Sciences Offenburg
Offenburg, Germany
axel.sikora@hs-offenburg.de

*Abstract—* *Training deep neural networks using backpropagation is very memory and computationally intensive. This makes it difficult to run on-device learning or fine-tune neural networks on tiny, embedded devices such as low-power micro-controller units (MCUs). Sparse backpropagation algorithms try to reduce the computational load of on-device learning by training only a subset of the weights and biases. Existing approaches use a static number of weights to train. A poor choice of this so-called backpropagation ratio limits either the computational gain or can lead to severe accuracy losses. In this paper we present TinyProp, the first sparse backpropagation method that dynamically adapts the back-propagation ratio during on-device training for each training step. TinyProp induces a small calculation overhead to sort the elements of the gradient, which does not significantly impact the computational gains. TinyProp works particularly well on fine-tuning trained networks on MCUs, which is a typical use case for embedded applications. For typical datasets from three datasets MNIST, DCASE2020 and CIFAR10, we are 5 times faster compared to non-sparse training with an accuracy loss of on average 1%. On average, TinyProp is 2.9 times faster than existing, static sparse backpropagation algorithms and the accuracy loss is reduced on average by 6 % compared to a typical static setting of the back-propagation ratio.*

*Keywords—sparse backpropagation, TinyML, Neural networks, efficient training*

## I. Introduction

Adding intelligence to an embedded device based on custom deep neural networks (DNN) promises a seamless experience tailored to the specific needs of each user while preserving the integrity of their personal data. The advantages of training the custom DNN on the device are [1]:

- **Data security:** Since no information needs to be shared with external entities, data security can be better ensured.
- **Latency:** Data transmission takes time and is often associated with a delay. When this process is eliminated, the result is available immediately.
- **Energy saving and cost:** If a DNN must be retrained in the cloud, there is a high overhead in energy costs for sending and receiving the DNN. however, if the DNN is only retrained, the energy consumption for communication may be higher than training the DNN locally.
- **Connection availability:** If the device relies on the internet to function and the connection to the internet is interrupted, data cannot be sent to the server. This for example happens, if you try to use a voice assistant and it does not respond, because it is not connected to the internet.

In contrast to cloud servers, embedded systems are mostly very resource limited. Typical limitations in embedded devices such as Micro-Controller Units (MCUs) are:

- **Limited device memory:** for on-device training, we need to save more than just the weights and biases, e.g. also the activations and gradients.
- **Limited energy resources**, e.g., due to being battery-powered: training requires complex calculations and, hence, more energy.
- **Limited computing power:** the device cannot run the training of the DNN in a given time.

There has been significant progress in deploying inference on such devices, which is often referred to as TinyML. Yet, training DNNs tends to be time-consuming [2], especially for DNNs with numerous trainable parameters. In order to be able to train DNNs on embedded devices, training algorithms have to be adapted to the limitations. For example, an important reason, why DNN training is typically slow, is that backpropagation requires the computation of the full gradient and updates all parameters in each learning step [3]. As DNN with many parameters become more prevalent, more efforts are being made to speed up the process of backpropagation.

Recent research has attempted to solve the problem of limited resources by demonstrating effective training on embedded devices. One technique is to approximate the actual gradient calculation itself [4, 3, 5, 6]. Another recent work [7,

8] also suggested that the exact gradient may not be necessary for efficient training of DNNs. Studies have shown that only the sign of the gradient is necessary for efficient backpropagation. Surprisingly, even random gradients can be used for efficient DNN training [9, 10]. However, these findings are mostly limited to small, fully connected networks on smaller datasets.

Since up to 90% of the computational time is spent performing these dot product operations [3], we focus on reducing their computational cost in this paper. We propose the sparse backpropagation technique TinyProp for DNN learning, where only a small subset of the full gradient is computed to update the corresponding trainable DNN parameters. In contrast to previously proposed sparse backpropagation algorithms, our TinyProp approach is adaptive. As a main contribution of this paper, we propose a method that computes how many elements of the gradient should be computed individually for each layer based on how well the model is already trained. In practice, it is usually the case that the DNN is already trained and only needs to be adapted for the user. With our approach, only as much training as necessary is done, regardless of whether the DNN is trained from scratch or fine-tuned. We show that our algorithm works by adapting the algorithm to different use cases and network architectures. For typical datasets from two datasets MNIST and DCASE2020, we have used less than 20% of the computational budget compared to non-sparse training. TinyProp achieves approximately the same results, about 1% less accuracy, as non-sparse training. This shows how useful sparse backpropagation can be. In addition, we have compared our method with existing sparse backpropagation methods with different top-k, and across all experiments we are faster and more accurate than the existing methods. On average, we are 2.9 times faster than the fixed top-k with the same accuracy and 6% more accurate than the fixed top-k with the same acceleration.

II. RELATED WORK

Training DNNs on embedded devices, and specifically MCUs, is a quite recent area of research. Most of the existing frameworks such as Tensorflow Micro [11], Edge-ML [12], Open-NN [13], etc., do not yet provide methods for on-device training.

To achieve resource-efficient training, researchers have so far focused to optimize existing algorithms for various resource-constrained systems. We can divide previous papers into three groups: Papers that implement other machine learning algorithm on embedded devices, papers that implement DNNs on embedded devices and papers that present a sparse backpropagation variation. First to the first group: Lee et al. [14] implemented a Gaussian mixture DNN on an embedded board with the aim of retraining an edge-level ML algorithm, which is very interesting, but TinyProp deals with the training of neural networks. SEFR [15], a low-power classifier, is the most related work training and inferring a binary inference on MCUs, so like [14] it is not aiming on neural networks. The same is true for MLMCU [16], which presents an optimization for the training process of linear models. Sudharsan et al. also presented Edge2Train [17], which optimizes SVMs for use on embedded devices.

Cartesiam NanoEdge AI Studio [18] allows the creation of static ML libraries to be embedded in CortexM MCUs. It enables the integration of the training process in the constrained device. Moreover, it can also train unsupervised algorithms on MCUs. However, the above work and other similar algorithms [19, 20] are tailored to specific applications and do not enable MCU-based IoT edge devices to learn/train themselves from a wide range of IoT application data.

The second group of papers that focus on training DNNs on embedded devices are the following:

TinyOL [1] is a technique to retrain the last layer of a DNN to react to changing environments. Although this technique is interesting for finetuning, it is unfortunately only limited to finetuning and does not cover the full training of a network. Furthermore, this technique can be covered with the help of TinyProp. This is similar for TinyTL [21]. Cai et al. present a method that only changes the bias of a DNN. Again, it is interesting for finetuning a model, but not for training from scratch. The Artificial Intelligence for Embedded Systems (AIfES) [22] library is a C-based, platform-independent tool for generating NNs that is compatible with a number of open-source MCU boards. AIfES can be used with Windows and embedded Linux platforms by providing efficient code in the form of a Dynamic Link Library (DLL). Unfortunately, AIfES does not use any efficient training algorithm, nevertheless TinyProp uses AIfES as a codebase.

The third research direction is to accelerate backpropagation with sparse backpropagation [3, 4, 23, 24], which aims to sparsify the full gradient vector to achieve significant computational cost savings. An effective solution to sparse backpropagation is top-k sparsity, which keeps only k elements with the largest absolute values in the gradient vector and backpropagates them across different layers. For example, meProp [3] uses top-k sparseness to compute only a very small but critical part of the gradient information and updates the corresponding DNN parameters. Going one step further, [4] implements top-k sparseness for backpropagation of convolutional neural networks. Experimental results show that these methods can significantly speed-up the backpropagation process. However, despite the success in saving computational cost, top-k sparsity for backpropagation still suffers from some hard-to-fix drawbacks, which are explained in more detail in the following section. [24] goes an interesting step further by determining the k differently. The authors do not assume a normal distribution in the gradients, but a log-normal distribution. This changes the determination of how high the k should be. Based on these techniques, we have developed TinyProp. In contrast to the other techniques, we do not assume a fixed top-k during training, but a dynamic top-k, which decreases over the training time.

**1. Forward Propagation**

$z^l = W^l \cdot a^l + b^l$
$a^{l+1} = f(z^l)$

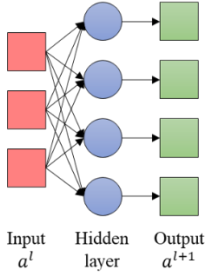

Input  Hidden  Output
$a^l$   layer   $a^{l+1}$

**2. Sum magnitudes of local error**

$Y^l = \sum_{i=1}^{N^l} |\delta_{a,i}^l|$

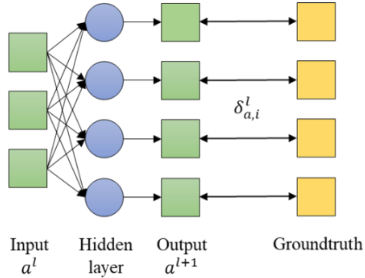

Input  Hidden  Output  Groundtruth
$a^l$   layer   $a^{l+1}$

**3. Calculate k**

$S^l := \left(S_{min} + Y^l \cdot \dfrac{S_{max} - S_{min}}{Y_{max}^l}\right) \cdot \zeta^{L-l}$

$k^l = S^l \cdot N^l$

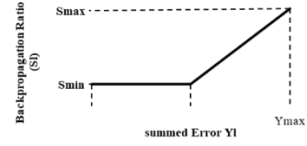

**4. Get Top k**
(Top k = 2)

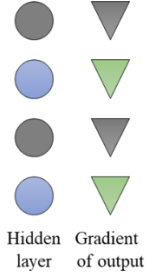

Hidden  Gradient
layer   of output

**5. Sparse back Propagation**
(Top k = 2)

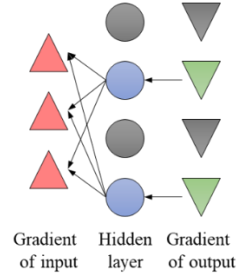

Gradient  Hidden  Gradient
of input  layer   of output

*Fig. 1. The steps that are carried out with TinyProp. 1. the forward pass is performed. 2. the loss function is calculated, and the local error is summed up 3. the local k for this layer is determined from the summed error. 4. the top k gradients are searched for. 5. the sparse backpropagation algorithm is performed.*

## III. TINYPROP METHOD

### A. Background: Sparse Backpropagation

In the case of feedforward DNNs, the operations that must be performed per layer during a training iteration are as follows:

$$z^l = W^l \cdot a^l + b^l$$
$$a^{l+1} = f(z^l) \quad (1)$$

with $W, b, z$ and $a$ being the weight tensor, bias, preactivation and activation values for layer $l$ respectively.

The training process is carried out using the backpropagation algorithm. First, the loss $\mathcal{L}$ is calculated:

$$\mathcal{L}(a^L, y) \quad (2)$$

with y being the ground truth. Then the backpropagation algorithm is started by calculating the local error for the last layer directly from the loss $\mathcal{L}$:

$$\delta_z^L = \left(\dfrac{\partial \mathcal{L}}{\partial z^L}\right)^T = \left(\dfrac{\partial \mathcal{L}}{\partial a^L}\dfrac{\partial a^L}{\partial z^L}\right)^T$$
$$= \left(\dfrac{\partial a^L}{\partial z^L}\right)^T \left(\dfrac{\partial \mathcal{L}}{\partial a^L}\right)^T = f'(z^L) \odot \nabla_{a^L}\mathcal{L} \quad (3)$$

and for all following layers l the gradients compute as follows:

$$\delta_z^l = \delta_a^l \odot f'(z^l) \quad (4)$$

$$\delta_a^{l-1} = (W^{l-1})^T \cdot \delta_z^l$$

$$\nabla_{W^l}\mathcal{L} = \delta_W^l = \delta_z^l \cdot (a^{l-1})^T \quad (5)$$

$\delta_W, \delta_b, \delta_z$ and $\delta_a$ denote the error or gradients of the respective quantities. With $f$ we denote the non-linear activation function whereas with $f'$ its derivative and $T$ denotes the transpose operation. Finally, the symbols $\cdot$ and $\odot$ denote the dot and Hadamard product respectively. As can be seen, three large matrix multiplications are required during a learning step on each layer, namely one in the forward pass in Eq. (1) and two in the backward pass (Eq. (4) and Eq. (5)).

The sparse backpropagation algorithm uses an approximated gradient by keeping only the top-k elements (based on magnitude) of the gradient $\delta_a^l$ and passing them through the gradient computation graph according to the chain rule. We denote the indices of the top k-values of the gradient vector as $\{t_1, t_2, ..., t_k\}$ $(1 \leq k \leq n)$, and the approximate gradient can then be expressed as:

$$\hat{\delta}_a^l \leftarrow \delta_a^l \ if \ a \ \epsilon \{t_1, t_2, ..., t_k\} \ else \ 0 \quad (6)$$

For example, suppose a vector $v = [1,2,3,-4]^T$, then $top(v, 2) = [0,0,3,-4]^T$. To approximate the original gradient $\delta_a^{l-1}$ and the weight gradient $\delta_W^l$, we now pass $\hat{\delta}_a^l$ through the gradient computation graph. As a result, the necessary matrix multiplications are performed in a sparsified manner, which leads to a linear reduction ($k$ divided by the respective vector dimension) of the computational effort. Consequently, the gradient of $W$ approaches to $\hat{W}$:

$$\nabla_{\hat{W}^l} \mathcal{L} = \hat{\delta}_W^l = \hat{\delta}_z^l \cdot (a^{l-1})^T \quad (7)$$

By doing so, only k rows or columns (depending on the layout) of the weight gradient are changed in each backwards pass. The formulas for computing the original gradient also change accordingly:

$$\begin{aligned} \hat{\delta}_a^l &= top(\delta_a^l, k) \\ \hat{\delta}_z^l &= \hat{\delta}_a^l \odot f'(z^l) \\ \widehat{\delta_a^{l-1}} &= (W^{l-1})^T \cdot \hat{\delta}_z^l \end{aligned} \quad (8)$$

Despite also using $\hat{\delta}_a^l$ to calculate the original gradient in a sparse matrix multiplication, the resulting vector remains fully occupied. This way the process can be repeated in the previous layer without getting more and more sparse gradients $\delta_a^l$.

### B. TinyProp Adaptivity Approach

The novelty of the TinyProp approach is to determine an adaptive $k$. A value of $k^l$ is determined individually and adaptively for each training input and each layer $l$. The adaptivity is that at TinyProp we calculate how many elements of the gradient of the trainable parameters of a layer should be calculated, called $S^l$, and $S$ is calculated by the local layer error. For example a value of $S^2 = 0.4$ indicates that 40% of the elements of the gradient of the trainable parameters of layer 2 should be computed for this sparse backpropagation step. An overview is shown in Fig. 1.

The basis for determining $k^l$ is the local error vector $\delta_{a,i}^l$ of a layer. In order to represent the entire layer, the sums of the individual $N^l$ components are first added up to obtain a total error $Y^l$ characteristic of the layer.

$$Y^l = \sum_{i=1}^{N^l} |\delta_{a,i}^l| \quad (9)$$

Our approach is to store a maximum value $Y_{max}^l$ of the layer error over the whole training process and, if necessary, update it during training to use it as a normalization. During ongoing training, this shift error reduces over time and thus allows a determination of $S^l$ that reflects the training progress. This normalization is independent of the number of neurons and is, therefore, a *general* approach for any kind of DNNs. The user has the possibility to define a maximum and minimum error propagation rate:

$$S_{max} \in [S_{min}, 1], S_{min} \in [0, S_{max}]$$

The error propagation rate can then be linearly interpolated between these values:

$$\hat{S}^l := S_{min} + Y^l \cdot \frac{S_{max} - S_{min}}{Y_{max}^l} \quad (10)$$

We use this function to solve a fundamental problem: The error propagation rate $S^l$ is defined exclusively in the interval [0,1], while the total error is not restricted as it is calculated from the DNN parameters. Although these are usually quite small, an upper limit cannot be guaranteed. Therefore, the total error, even if divided by the respective number of neurons in the layer, remains unbound. No maximum total error of a stratum can be determined that could be assigned to a rate of $S^l = 1$.

### C. TinyProp Damping Factor

The largest DNN layers are usually the first ones. While these have the largest matrices, they need little or no training for fine-tuning, because features usually do not change. However, due to the layer-specific normalization of the total error as described above, front layers would have similarly high error propagation rates as, for example, the last DNN layer. Therefore, a lot of time is spent here with calculations that are ultimately not needed.

For this reason, another hyperparameter is introduced, which is additionally applied to $S^l$ for each layer. This also allows TinyProp to provide a similar function as TinyOL [1], If the DNN to be considered has $L$ layers and $l$ describes the layer index with the bounds $1 \leq l \leq L$, this layer-dependent damping factor $\zeta(l)$ damps the error propagation rate $S$ for each layer.

$$S^l = \hat{S}^l \cdot \zeta^{-l+L} \quad (11)$$

For example, if one defines an attenuation factor of $\zeta = 0.5$, the last error propagation rate $S^l$ remains unaffected, while that of the penultimate layer $S^{l-1}$ is halved. Its antecedent $S^{l-2}$ is in turn only a quarter of its original value. Substituting the interpolated value for S from equation 10 into equation 11 gives the complete equation for determining S in a layer.

With $S^l$ the top-k can then be calculated adaptively:

$$k^l = S^l \cdot N^l \quad (12)$$

Combining all steps, we get the TinyProp backpropagation algorithm:

$$\begin{aligned} S^l &= \left( S_{min} + Y^l \cdot \frac{S_{max} - S_{min}}{Y_{max}^l} \right) \cdot \zeta^{-l+L} \\ k^l &= S^l \cdot N^l \\ \hat{\delta}_a^l &= top(\delta_a^l, k) \\ \widehat{\delta_z^l} &= \widehat{\delta_a^l} \odot f'(z^l) \\ \widehat{\delta_a^{l-1}} &= (W^{l-1})^T \cdot \hat{\delta}_z^l \\ \widehat{\delta_W^l} &= \widehat{\delta_z^l} \cdot (a^{l-1})^T \end{aligned} \quad (13)$$

The implementation of the algorithm was done in C. We decided to develop the method based on the open-source framework AIfES [22] to be able to guarantee its use in the tinyML area. AIfES is a platform-independent and constantly growing machine learning library in the programming language C, which exclusively uses standard libraries based on the GNU Compiler Collection (GCC).

With the above procedure, a modified version of the backpropagation can now be created that includes the computation of $k^l$ as described in the previous section. This function is called per layer in each backward pass. The time saved by the sparse matrix multiplications clearly outweighs the additional computational effort by determining the total error and the top-k largest gradients, which we were able to show through the results.

The function sequence is illustrated in the following pseudocode:

```
//sum magnitudes of local error
   for every neurons as i
total_error = sum absolute(layer_error[i]);
//record max total error
if total_error greater then max_error
  max_error = total_error;
interpolate S;
dampen S;
k = S*number_of_neurons
//selection method
get top k indices;
calculate sparse weight gradient;
calculate bias gradient;
calculate sparse error of previous layer;
```

## IV. EXPERIMENTS

To demonstrate the benefit of the proposed method, we tested it on different open-source datasets:

### A. MNIST

We use the MNIST dataset for handwritten digits [25] for a first evaluation. MNIST consists of 60,000 28×28-pixel training images and an additional 10,000 test samples. Each image contains a single numeric digit (0-9). The neural network we used was a multilayer dense net.

### B. Anomalie

For this benchmark, we chose to use the DCASE2020 [26] competition dataset. The DCASE dataset contains data for six machine types (slide rail, fan, pump, valve, toy car, toy conveyor), and the competition rules allowed the training of separate DNNs for each model. We took for our experiments only the slide rail machine type. For the training, normal sound samples of seven different slide rails are provided, each containing on average 1024 samples mixing the machine noise with ambient noise. To transform the problem into a supervised learning problem, we have divided the samples into two classes: Anomaly and no anomaly. The neural network we used was a multilayer dense net.

### C. CIFAR10

To show that the method also works successfully with complex data sets, we use the CIFAR10 [27] data set as a benchmark. The CIFAR-10 dataset consists of 60000 32x32 color images in 10 classes, with 6000 images per class. There are 50000 training images and 10000 test images. For this we used a 2d convolutional neural network.

### D. Experimental Settings

Two types of experiments were conducted to demonstrate the practical benefits of TinyProp:

1) Training from scratch. The DNN was reinitialized and trained from scratch.
2) Fine-tuning of a pre-trained DNN. The DNN was pre- trained with the normal backpropagation algorithm until an accuracy of 85% on MNIST and Anomaly and 75% on CIFAR10 on test data and then re-trained with the sparse backpropagation algorithm.

In addition to the baseline, which is trained with the normal backpropagation algorithm, other experiments were also carried out with fixed k to be able to compare TinyProp with the other works [3, 4] In order to be able to compare TinyProp better, we have converted the fixed k into a backpropagation ratio. This means that for a network with, for example, 1000 weights, we have only updated 333 weights with a calculated backpropagation ratio of 33%. Table I shows the fixed k and the converted backpropagation ratio.

Thus, in addition to the presented technique, the experiments were carried out with a backpropagation ratio of 100% (baseline), 10%, 15%, 33% and 66%.

The experiments were carried out on 1 core of a ESP32 with 240 MHz.

To ensure that the comparison of our technique with the state-of-the-art methods and the baseline is consistent and logical for the use case in the TinyML area, we leave the number of epochs the same for each data set. This means that the baseline, the experiments with fixed k and our method were trained with the same number of epochs. To give a factor of time savings, we have averaged the times it takes the algorithm to perform an epoch. Many works in this area compare only the time of the backpropagation step, but neglect the time needed to sort the k. By averaging the time of the entire epochs, all steps are included, and thus the time saved in which the MCU is busy is given.

### E. Experimental Results

1) *Training from scratch:*

In the experiments we conducted, we looked at the results in terms of accuracy, the total computational operations required and the speed-up of the average training step. With the experimental setup we tried to test different DNN architectures, different data types and different hyperparameter settings on our algorithm. The results, which can be seen in Table I, are average values of 100 runs with randomized initializations. In the results, several special features of TinyProp can be seen, starting with the observation that TinyProp is faster than the baseline by an acceleration factor up to 4.4 times, with an accuracy loss of only 0.3%. Furthermore, TinyProp also delivers better results than sparse backpropagation algorithms with a fixed k. While fixed sparse backpropagation algorithms quickly lose accuracy as the backpropagation ratio decreases, TinyProp can still deliver good training results with only 18% of backpropagation ratio.

TABLE I
RESULTS OF THE COMPARISON BETWEEN TINYPROP AND SPARSE BACKPROPAGATION ALGORITHM WITH FIXED K

| MNIST scratch | Baseline | top-k 6500 | top-k 12000 | top-k 17000 | top-k 30000 | top-k 66000 | TinyProp |
|---|---|---|---|---|---|---|---|
| Accuracy (%) | 96.6 | 21.2 | 40.9 | 86.3 | 90.2 | 92.1 | 96.3 |
| Back propagation Ratio | 1 | 0.1 | 0.15 | 0.2 | 0.33 | 0.66 | 0.18 |
| Runtime ESP32 per Epoch | 150.15s | 25.024s | 30.03s | 37.51s | 50s | 100.1s | 34,125s |
| Acceleration | 1x | 6x | 5x | 4x | 3x | 1.5x | 4,4x |
| DCASE2020 scratch | Baseline | top-k 30000 | top-k 65000 | top-k 100000 | top-k 190000 | top-k 430000 | TinyProp |
| Accuracy (%) | 98.88 | 50.01 | 85.7 | 93.68 | 94.44 | 96.11 | 96.75 |
| Back propagation Ratio | 1 | 0.1 | 0.15 | 0.2 | 0.33 | 0.66 | 0.15 |
| Runtime ESP32 per Epoch | 72s | 12.01s | 14.45s | 18s | 23.97s | 47.99s | 16.36s |
| Acceleration | 1x | 6x | 5x | 4x | 3x | 1.5x | 5x |
| CIFAR10 scratch | Baseline | top-k 30000 | top-k 65000 | top-k 100000 | top-k 190000 | top-k 430000 | TinyProp |
| Accuracy (%) | 78.9 | 16.8 | 24.9 | 53.5 | 67.1 | 70.2 | 77.1 |
| Back propagation Ratio | 1 | 0.1 | 0.15 | 0.2 | 0.33 | 0.66 | 0.19 |
| Runtime ESP32 per Epoch | 315s | 52.5s | 63s | 78.75s | 105s | 210s | 77s |
| Acceleration | 1x | 6x | 5x | 4x | 3x | 1.5x | 4x |
| MNIST fine-tuning | Baseline | top-k 6500 | top-k 12000 | top-k 17000 | top-k 30000 | top-k 66000 | TinyProp |
| Accuracy (%) | 96.4 | 85.2 | 85.9 | 86.0 | 89.9 | 91.9 | 96.1 |
| Back propagation Ratio | 1 | 0.1 | 0.15 | 0.2 | 0.33 | 0.66 | 0.07 |
| Runtime ESP32 per Epoch | 150.15s | 25.024s | 30.03s | 37.51s | 50s | 100.1s | 18,1s |
| Acceleration | 1x | 6x | 5x | 4x | 3x | 1.5x | 8,3x |
| DCASE2020 fine-tuning | Baseline | top-k 30000 | top-k 65000 | top-k 100000 | top-k 190000 | top-k 430000 | TinyProp |
| Accuracy (%) | 98.4 | 85.1 | 85.1 | 93.2 | 94.2 | 96.9 | 97.55 |
| Back propagation Ratio | 1 | 0.1 | 0.15 | 0.2 | 0.33 | 0.66 | 0.05 |
| Runtime ESP32 per Epoch | 72s | 12.01s | 14.45s | 18s | 23.97s | 47.99s | 7.21s |
| Acceleration | 1x | 6x | 5x | 4x | 3x | 1.5x | 10x |
| CIFAR fine-tuning | Baseline | top-k 90000 | top-k 135000 | top-k 180000 | top-k 297000 | top-k 900000 | TinyProp |
| Accuracy (%) | 78.6 | 75.1 | 75.2 | 75.5 | 76.9 | 77.2 | 77.9 |
| Back propagation Ratio | 1 | 0.1 | 0.15 | 0.2 | 0.33 | 0.66 | 0.1 |
| Runtime ESP32 per Epoch | 315s | 52.5s | 63s | 78.75s | 105s | 210s | 52.7s |
| Acceleration | 1x | 6x | 5x | 4x | 3x | 1.5x | 6x |

A similar picture emerges with the Anomaly Detection dataset. Again, TinyProp has only a minimal loss of accuracy of 2.13% compared to the baseline and this with a backpropagation ratio of 15% of the entire network. The acceleration with this backpropagation ratio is up to 5 times faster than train the model full. Compared to a fixed k, TinyProp is always better in accuracy and in the experiments where a fixed k is faster, the accuracies with a fixed k are too poor for the results to be meaningful.

The model with the complex data set CIFAR10 can also be trained efficiently with TinyProp. With the help of TinyProp, the mesh is trained 4 times faster at a cost of only 1.8% points. On average, only 19% of the net is trained. This means that TinyProp again outperforms every top-k variant.

*2) Fine-tuning with TinyProp*

The results for another practical use case of fine-tuning in the TinyML domain clearly show the strength of a dynamic sparse algorithm.

The newly introduced hyperparameters $S_{max}, S_{min}$ and $\zeta$ make it easy to determine how many resources are available. In this way, the training algorithm can be adapted whether a network is trained from scratch or, as here, the network is only trained after.

The second half of Tables I show that TinyProp is again superior to a fixed top-k and can go even further down with the backpropagation rate, in contrast to training from scratch. over the entire training process, only 7% of the network was trained with MNIST and only 5% with the Anomaly Detection dataset and CIFAR10 was only 10% trained. In every use case we again have a loss of accuracy of less than 1%. This is also reflected in the acceleration results, across each experiment TinyProp is faster. In contrast to the full training, TinyProp is 8.3 times faster in the MNIST case, 10 times faster for the anomaly detection and 6 times faster for CIFAR10.

## V. Hyperparameter Selection

We have introduced new hyperparameters with TinyProp to better control the available resources of the embedded systems in the training. In this chapter, we would like to explain how the hyperparameters should be chosen to adapt the algorithm to the target device in the best possible way.

We have conducted several experiments with different sets of hyperparameters. Each of these experiments was performed 100 times with random initializations.

The result of the experiments was that the exact settings are not that important to get a good result. However, the more we allow to train, the better the result. The most important hyperparameter is $S_{max}$. With the help of $S_{max}$, we can control whether a net is trained from the scratch or whether the net is fine-tuned.

A recommendation drawn from the experience of the experiments is as follows:

TABLE II
TINYPROP HYPERPARAMETER

|  | $S_{max}$ | $S_{min}$ | $\zeta$ |
| --- | --- | --- | --- |
| Training from scratch | 0.8 | 0.1 | 0.9 |
| Fine-Tune DNN | 0.4 | 0.1 | 0.9 |

## VI. Conclusion

The backpropagation algorithm attempts to improve all parameters in the training step simultaneously. However, previous work has shown that this approach could be more efficient. With TinyProp, we have further developed the techniques of sparse backpropagation with an adaptive factor. The dynamic factor consists of deciding separately for each training data point how much to train the network. We were able to show that DNNs can be trained faster and better using this dynamic factor than if no dynamic factor is allowed in sparse training.

TinyProp makes sparse training even more efficient and trains only what needs to be trained to improve the outcome. In addition, we were able to show that our approach can halve the training time compared to full training, with minimal loss of accuracy.

[1] Haoyu Ren, Darko Anicic, and Thomas Runkler. TinyOL: TinyML with Online-Learning on Microcontrollers.
[2] Sébastien Jean et al. On Using Very Large Target Vocabulary for Neural Machine Translation.
[3] Xu Sun et al. meProp: Sparsified Back Propagation for Accelerated Deep Learning with Reduced Overfitting.
[4] Bingzhen Wei et al. Minimal Effort Back Propagation for Convolutional Neural Networks.
[5] Xu Sun et al. Training Simplification and Model Simplification for Deep Learning: A Minimal Effort Back Propagation Method. 2020. DOI: 10.1109/TKDE.2018.2883613.
[6] Menachem Adelman et al. Faster Neural Network Training with Approximate Tensor Operations.
[7] Wei Wen et al. TernGrad: Ternary Gradients to Reduce Communication in Distributed Deep Learning.
[8] Will Xiao et al. Biologically-plausible learning algorithms can scale to large datasets.
[9] Timothy P. Lillicrap et al. "Random synaptic feedback weights support error backpropagation for deep learning". In: *Nature communications* 7 (2016), p. 13276. DOI: 10.1038/ncomms13276.
[10] Arild Nøkland. Direct Feedback Alignment Provides Learning in Deep Neural Networks.
[11] TensorFlow. *TensorFlow Lite — ML for Mobile and Edge Devices*. 11.12.2021. URL: https://www.tensorflow.org/lite/.
[12] GitHub. microsoft/EdgeML: This repository provides code for machine learning algorithms for edge devices developed at Microsoft Research India. 1.02.2022. URL: https://github.com/Microsoft/EdgeML.
[13] OpenNN — Open Neural Networks Library. 23.11.2021. URL: https://www.opennn.net/.
[14] Jongmin Lee et al. "Integrating machine learning in embedded sensor systems for Internet-of-Things applications". In: *2016 IEEE International Symposium on Signal Processing and Information Technology (ISSPIT)*. IEEE, 12/12/2016 - 12/14/2016, pp. 290–294. DOI: 10.1109/ISSPIT.2016.7886051.
[15] Hamidreza Keshavarz, Mohammad Saniee Abadeh, and Reza Rawassizadeh. *SEFR: A Fast Linear-Time Classifier for Ultra-Low Power Devices*.
[16] Bharath Sudharsan, John G. Breslin, and Muhammad Intizar Ali. "ML-MCU: A Framework to Train ML Classifiers on MCU-based IoT Edge Devices". In: *IEEE Internet of Things Journal* (2021), p. 1. DOI: 10.1109/JIOT.2021.3098166.
[17] Bharath Sudharsan, John G. Breslin, and Muhammad Intizar Ali. "Edge2Train". In: *Proceedings of the 10th International Conference on the Internet of Things*. Ed. by Paul Davidsson and Marc Langheinrich. New York, NY, USA: ACM, 10062020, pp. 1–8. DOI: 10.1145/3410992.3411014.
[18] Cartesiam. Cartesiam, NanoEdge AI Library. 1.12.2021. URL: https://cartesiam.ai/.
[19] Goutham Kamath et al. "Pushing Analytics to the Edge". In: 2016 IEEE Global Communications Conference (GLOBECOM). IEEE, 12/4/2016 - 12/8/2016, pp. 1–6. DOI: 10.1109/GLOCOM.2016.7842181.
[20] Ashish Kumar, Saurabh Goyal, and Manik Varma. "Resource-Efficient Machine Learning in 2 KB RAM for the Internet of Things". In: *Proceedings of the 34th International Conference on Machine Learning - Volume 70*. ICML'17. Sydney, NSW, Australia: JMLR.org, 2017, pp. 1935–1944.
[21] Han Cai et al. *TinyTL: Reduce Activations, Not Trainable Parameters for Efficient On-Device Learning*.
[22] GitHub. Fraunhofer-IMS/AIfES for Arduino: This is the Arduino® compatible port of the AIfES machine learning framework, developed and maintained by Fraunhofer Institute for Microelectronic Circuits and Systems. 7.02.2022. URL: https://github.com/Fraunhofer-IMS/AIfES_for_Arduino.
[23] Maohua Zhu et al. *Structurally Sparsified Backward Propagation for Faster Long Short-Term Memory Training*.
[24] Brian Chmiel et al. Neural gradients are near-lognormal: improved quantized and sparse training. URL: https://arxiv.org/pdf/2006.08173.
[25] LeCun et al. (1999): The MNIST Dataset Of Handwritten Digits (Images) — PyMVPA 2.6.5.dev1 documentation. 9.04.2019. URL: http://www.pymvpa.org/datadb/mnist.html.
[26] Dcase. *DCASE2020 Challenge - DCASE*. 1.02.2022. URL: http://dcase.community/challenge2020/index.
[27] CIFAR-10 and CIFAR-100 datasets. (n.d.). Retrieved November 19, 2022, from https://www.cs.toronto.edu/~kriz/cifar.html